\documentclass[conference]{IEEEtran}
\usepackage{amsmath,amssymb,amsfonts}
\usepackage{algorithmic}
\usepackage{graphicx}
\usepackage{textcomp}
\usepackage{color}
\usepackage{multirow}
\usepackage{hhline}

\usepackage{subcaption}

\ifCLASSINFOpdf
\else
\fi
\begin{document}

\title{Regularizing Autoencoder-Based Matrix Completion Models via Manifold Learning}
%\author{Duc Minh Nguyen, Evaggelia~Tsiligianni, Robert Calderbank, and~Nikos Deligiannis% <-this % stops a space
%\thanks{D.~M.~Nguyen, E.~Tsiligianni and~N.~Deligiannis are with ETRO, Vrije Universiteit Brussel, Pleinlaan 2, B-1050 Brussels, Belgium, and imec, Kapeldreef 75, B-3001 Leuven, Belgium. E-mail:\{mdnguyen, etsiligi, ndeligia\}@etrovub.be. 
%R.~Calderbank is with ECE, Duke University, Durham, North Carolina.  E-mail: robert.calderbank@duke.edu. }
%}

\author{Duc Minh Nguyen$^{\star}$,
Evaggelia~Tsiligianni$^{\star}$,
        Robert Calderbank$^{\dagger}$,
        Nikos Deligiannis$^{\star}$,% <-this % stops a space
        \\$^\star$ Department of Electronics and Informatics, Vrije Universiteit Brussel, Pleinlaan 2, B-1050 Brussels, Belgium \\
                          imec, Kapeldreef 75, B-3001 Leuven, Belgium \\
                          %\{mdnguyen, etsiligi, ndeligia\}@etrovub.be
        $^\dagger$ Department of Electrical and Computer Engineering, Duke University, Durham, North Carolina 
        	%robert.calderbank@duke.edu% 
        }    

\maketitle

\begin{abstract}
Autoencoders are popular among neural-network-based matrix completion models
due to their ability to retrieve potential latent factors from the partially observed matrices.
Nevertheless, when training data is scarce their performance is significantly degraded due to overfitting.
In this paper, we mitigate overfitting with a data-dependent regularization technique
that relies on the principles of multi-task learning.
Specifically, we propose an autoencoder-based matrix completion model 
that performs prediction of the unknown matrix values as a main task,  
and manifold learning as an auxiliary task.
The latter acts as an inductive bias, 
leading to solutions that generalize better.
The proposed model outperforms the existing autoencoder-based models designed for matrix completion,
achieving high reconstruction accuracy in well-known datasets.
\end{abstract}


\section{Introduction}
Recovering the unknown entries of a partially observed matrix
is an important problem in signal processing~\cite{Cao2014, Ji2010} and machine learning~\cite{zheng16, monti17}. 
This prblem is often referred to as matrix completion (MC). 
Let $M \in \mathbb{R}^{n \times m}$ be a matrix with a limited number of observed entries
defined by a set of indices $\Omega$ such that $(i,j) \in\Omega$ if $M_{ij}$ has been observed. 
Then, recovering $M$ from the observed entries can be formulated as an optimization problem of the form:
\begin{equation}
        R^\star = \arg\min_ {R}\|P_{\Omega}(R-M)\|_F,
\label{eq:mc}
\end{equation}
with $R^\star \in \mathbb{R}^{n \times m}$ denoting the complete matrix,
$P_{\Omega}$ an operator that indexes the entries defined in $\Omega$,
and $\|\cdot \|_F$ the Frobenius norm, that is,
$\| {P}_{\Omega}(R-M) \|_F = \big( \sum_{(i,j) \in \Omega}(R_{ij}-M_{ij})^2 \big)^{1/2}$.

Several studies have focused on the problem of reconstructing $R^\star$ from $M$,
most of them assuming that $R^\star$ is a low-rank matrix.
As the rank minimization problem is intractable, 
nuclear norm minimization is often used as a convex relaxation~\cite{candes09, Recht10}.
However, the major drawback of nuclear norm minimization algorithms is their high computational cost. 
%especially when the dimensions of the matrix increase.
Less computationally demanding matrix factorization methods~\cite{salakhutdinov07, Koren2009} 
approximate the unknown matrix by a product of two factors 
$U \in \mathbb{R}^{n \times r}$, $V  \in \mathbb{R}^{m \times r}$ with $r\ll \min(n,m)$,
and have been employed in large-scale MC problems
such as recommender systems~\cite{Koren2009}.

Recently, matrix completion has been addressed
by several deep-network-based models~\cite{zheng16, monti17, volkovs17, nguyen18, nguyen18_spl} 
that yield state-of-the-art results in a variety of benchmark datasets. 
Having the ability to provide powerful data representations,
deep neural networks have achieved great successes
in many problems, 
from acoustic modeling~\cite{mohamed12} and compressed sensing~\cite{nguyen17} to 
image classification~\cite{he16} and social media analysis~\cite{do18}.
Among deep-network-based MC models, 
autoencoder-based methods have received a lot of attention 
due to their superior performance and a direct connection to 
the matrix factorization principle~\cite{sedhain15, strub16, strub16b, dong17, kuchaiev17}. 
Despite their remarkable performance, 
these models suffer from overfitting, 
especially, when dealing with matrices that are highly sparse. 
To overcome this problem, more recent works either employ 
data-independent regularization techniques 
such as weight decay and dropout~\cite{sedhain15, strub16, strub16b, kuchaiev17, Dropout}, 
or incorporate side information to mitigate the lack of available information~\cite{strub16, strub16b, dong17}. 
The efficiency of the former approach is limited
under high sparsity settings~\cite{sedhain15, strub16, kuchaiev17}, 
while the latter is not directly applicable 
when side information is  unavailable or difficult to collect. 

In this paper, we focus on matrix completion without side information 
under settings with highly scarce observations,
and propose a data-dependent regularization technique 
to mitigate the overfitting of autoencoder-based MC models. 
While data-independent regularization approaches 
focus on training a model such that it is difficult to fit to random error or noise,
data-dependent techniques rely on the idea that 
the data of interest lie close to a manifold
and learn attributes that are present in the data~\cite{Belkin06, GraphConnect}.
In particular, we combine row-based and column-based autoencoders 
into a hybrid model and constrain the latent representations produced by the two models 
by a manifold learning objective as in~\cite{nguyen18}. 
This constraint can be considered as a data-dependent regularization.
The resulting model follows the multi-task learning paradigm \cite{Caruana1997, Evgeniou05, ruder17}, 
where the \textit{main task} is the reconstruction 
and the \textit{auxiliary task} is the manifold learning.
Experimental results on various real datasets show that 
our model effectively mitigates overfitting 
and consistently improves the  performance of autoencoder-based models. 
The proposed approach is complementary to data-independent regularization,
and the two techniques can be combined in an efficient way~\cite{GraphConnect}.

The paper is organized as follows: 
Section \ref{sec:autorec} describes the general autoencoder-based MC model
and briefly reviews the literature in multi-task learning.
Section \ref{sec:proposed} presents the proposed model. 
Experimental results are presented in Section \ref{sec:experiment}, 
while Section \ref{sec:conclusion} concludes our paper.

\section{Background}
\label{sec:autorec}
%In this section, we first describe the autoencoder-based models in details, 
%as they are the focus of this paper. 
%After that, we briefly review the literature in multi-task learning. 
%
%------------------------------------------------------------------------------%
\subsection{Autoencoder-based Matrix Completion}
\begin{figure}
\centering
\vspace{0.1in}
\includegraphics[width=0.65\linewidth]{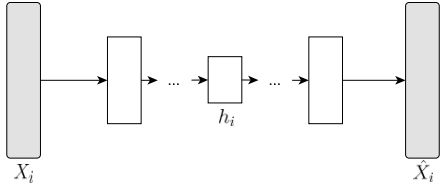}
\caption{The Autoencoder-based model in \cite{sedhain15}:   
$X_i$ is a partially observed row vector, corresponding to the $i$-th row in the original matrix. 
The model transforms $X_i$ to a hidden representation $h_i$ (encoding), 
and produces a dense reconstruction $\hat{X}_i$ from $h_i$ (decoding).  
}
\label{fig:autorec}
\end{figure}

%------------------------------------------------------------------------------%
The first autoencoder-based MC model, coined AutoRec~\cite{sedhain15},
comes in two versions, namely, the row-based and the column-based AutoRec.
For brevity, we only focus on the formulation of the row-based AutoRec; 
the column-based model can be defined in a similar way. 
The row-based AutoRec model operates on a row of a partially observed matrix, 
projects it into a low-dimensional latent (hidden) space and then reconstructs it 
in the outer space.
Let $M  \in \mathbb{R}^{n \times m}$ be an incomplete matrix 
with rows $X_i \in \mathbb{R}^m$,  $i=1\dots n$,
and columns $Y_j \in \mathbb{R}^n$, $j = 1\dots m$. 
In the encoder side, the model takes as input a partially observed row vector $X_i$ 
and transforms it to a latent representation vector $h_i$  
through a series of hidden neural network layers. 
In the decoder side,
a dense reconstruction of $X_i$ is generated from $h_i$ 
through another set of hidden layers
to predict the missing values. 
Denote by $f^e_{\scriptscriptstyle X}$ and $f^d_{\scriptscriptstyle X}$ the non-linear functions 
corresponding to the encoder and the decoder, respectively,
with $w_{\scriptscriptstyle X}^e$, $w_{\scriptscriptstyle X}^d$  
the corresponding vectors containing the free parameters of the model. 
Then, the intermediate representation of $X_i$ is given by $h_i = f^e_{\scriptscriptstyle X}(X_i; w_{\scriptscriptstyle X}^e)$,
while the dense reconstruction $\widehat{X}_i$ is obtained as 
$\widehat{X}_i = f^d_{\scriptscriptstyle X}(h_i; w_{\scriptscriptstyle X}^d) = f^d_{\scriptscriptstyle X}\big(f^e_{\scriptscriptstyle X}(X_i; w_{\scriptscriptstyle X}^e);w_{\scriptscriptstyle X}^d \big)$.
The objective function used to train the row-based AutoRec 
involves a reconstruction loss of the form:
\begin{equation}
\label{eq:autorecLoss}
	\mathcal{L}_{\scriptscriptstyle X} = \dfrac{1}{|\Omega|} \sum_{ij \in \Omega} \big(X_i(j) - \widehat{X}_i(j)\big)^2,
\end{equation}
and a regularization term $\mathcal{L}_\text{reg}(w_{\scriptscriptstyle X}^e,w_{\scriptscriptstyle X}^d)$;
typically, $\ell_2$ regularization is applied to the model parameters.
In the above equation,  $X_i(j)$ denotes the $j$-th element of the $i$-th row and
$|\Omega|$ the cardinality of  $\Omega$.
Figure \ref{fig:autorec} illustrates the general architecture of a row-based AutoRec. 
%
%The AutoRec model can be considered as performing \textbf{non-linear} matrix factorization \cite{strub16}. ...
%
Based on the same general architecture, 
more recent works~\cite{strub16, strub16b, kuchaiev17} 
propose improvements to the AutoRec model. 
We discuss these approaches in Sec.~\ref{sec:related work}
where we highlight the differences between existing work and the proposed model.
\subsection{Multi-task Learning}
When training a model designed for a specific task 
using information extracted by other tasks that are related to the main task,
we refer to multi-task learning~\cite{Caruana93, Caruana1997, Evgeniou05}.
The use of knowledge gained from the solution of one or more auxiliary tasks
is an inductive transfer mechanism, which can improve a model by introducing inductive bias.
The inductive bias provided by the auxiliary tasks causes the model to prefer 
hypotheses that explain more than one task.
This approach often leads to solutions that generalize better for the main task~\cite{ruder17}.
Multi-task learning can be considered, therefore, as a regularization technique
that reduces the risk of overfitting.
When designing a multi-task learning model,
we seek for auxiliary tasks that are similar to the main task.
Different approaches for considering two tasks as similar 
have been reported in the literature~\cite{Caruana1997, Baxter2000, BenDavid2003}.

Multi-task learning in deep neural networks
typically involves learning tasks in parallel while using a shared representation. 
A common approach when designing a multi-task deep network
is to share hidden layers between all tasks and separate task-specific outputs.
The idea is that what is learned for each task 
can help other tasks be learned better~\cite{ruder17}.
%
%
%----------------------------------------------------%
\section{Proposed Approach}
\label{sec:proposed}
%----------------------------------------------------%
Considering that overfitting is a major drawback of existing AutoRec models,
the MC model proposed in this paper aims to address this problem
by introducing a regularization strategy
that follows the multi-task learning principle.
The proposed model is a two-brach autoencoder performing one main task and one auxiliary task. 
The main task outputs row and column predictions of the unknown matrix.
The auxiliary task uses the latent representations of the main task
to predict the value of a single matrix entry.
Concretely, the additional output that corresponds to the auxiliary task  
imposes a similarity constraint on the hidden representations
produced by the row and column autoencoders.
The proposed constraint acts as  a data-dependent regularization 
to the autoencoder models.
\subsection{Model Architecture}
\begin{figure}
\centering
\vspace{0.1in}
\includegraphics[width=0.65\linewidth]{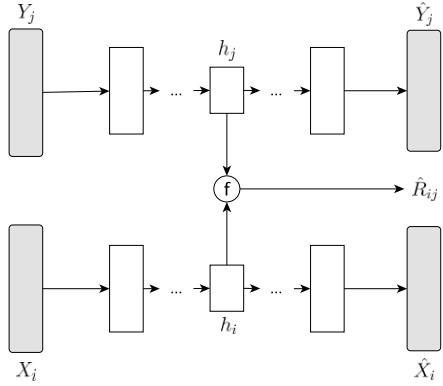}
\caption{The proposed model: column (top) and row (bottom) autoencoders 
produce dense column and row reconstructions ($\hat{Y}_j$ and $\hat{X}_i$). 
The column and row hidden representations $h_j$ and $h_i$ are constrained 
by a manifold learning objective.}
\label{fig:proposed}
\end{figure}

The architecture of the proposed model is illustrated in Fig.~\ref{fig:proposed}. 
The model consists of two branches. 
The row branch takes the $i$-th row vector $X_i$ as input,
predicts the missing values and produces the dense reconstruction $\widehat{X}_i$ as output.
Assuming a deep autoencoder with $L$ encoding and $L$ decoding layers,
the hidden representation of the $i$-th row is obtained according to:
\begin{align}
\label{eq:rowHidden}
h_i = & \sigma(W_{\scriptscriptstyle X(e)}^{(L)} a_{\scriptscriptstyle X(e)}^{(L-1)} + b_{\scriptscriptstyle X(e)}^{(L)}), \nonumber \\
a_{\scriptscriptstyle X(e)}^{(l)} = & \sigma (W_{\scriptscriptstyle X(e)}^{(l)} a_{\scriptscriptstyle X(e)}^{(l-1)} + b_{\scriptscriptstyle X(e)}^{(l)}), 
\ l = 1, \dots , L-1,
\end{align}
with $a_{\scriptscriptstyle X(e)}^{(0)} = X_i$,
while for the decoding part it holds
\begin{align}
\label{eq:rowAE}
\widehat{X}_i = & W_{\scriptscriptstyle X(d)}^{(L)} a_{\scriptscriptstyle X(d)}^{(L-1)} + b_{\scriptscriptstyle X(d)}^{(L)}, \nonumber \\
a_{\scriptscriptstyle X(d)}^{(l)} = & \sigma (W_{\scriptscriptstyle X(d)}^{(l)} a_{\scriptscriptstyle X(d)}^{(l-1)} + b_{\scriptscriptstyle X(d)}^{(l)}), 
\ l = 1, \dots , L-1,
\end{align}
with $a_{\scriptscriptstyle X(d)}^{(0)} = h_i$.
In the above equations, $W_{\scriptscriptstyle X(e)}^{(l)}$, $W_{\scriptscriptstyle X(d)}^{(l)}$ are the weights used 
in the $l$-th  encoding and the $l$-th decoding layer, respectively, 
$b_{\scriptscriptstyle X(e)}^{(l)}$, $ b_{\scriptscriptstyle X(d)}^{(l)}$
the corresponding biases,
and $\sigma(\cdot)$ the activation function.

Similarly, the column branch outputs the dense reconstruction $\widehat{Y}_j$ 
from the input incomplete column vector $Y_j$.
Without any additional constraint on these two branches, 
this model can be seen as an ensemble of two AutoRec models, working independently. 
To enforce that the two branches work together, 
we propose to use the corresponding latent representations of rows and columns
to perform a related task, namely, to realize a matrix factorization model.
The basic assumption of matrix factorization models is that 
there exists an unknown representation of rows and columns in a latent space
such that the $(i,j)$ entry of a matrix $M$ can be modelled as 
the inner product of the latent representations of 
the $i$-th row and the $j$-th column in that space~\cite{salakhutdinov07, Koren2009}.
This was the dominant idea of the deep matrix factorization model 
presented in~\cite{nguyen18}.
Following this approach,
we enforce the hidden representations of the row and column autoencoders
to be close under the cosine similarity 
if the corresponding matrix entry  is of high value.
Specifically, we define a loss term associated to the $(i,j)$ observed entry 
as follows:
\begin{equation}
\label{eq:reprLoss}
\mathcal{L}^\text{repr}_{ij} = \big(f(h_i, h_j) - M_{ij} \big)^2, \quad  (i,j) \in \Omega,
\end{equation}
where $f = \dfrac{h_{i}^{T} h_{j}}{\left\|h_i \right\|_2 \left\|h_j \right\|_2}$, is the cosine similarity function,  
and $h_i$, $h_j$  the hidden representations of the row and column autoencoders, respectively.
We employ the cosine similarity rather than 
the dot product as a similarity metric, 
as we empirically found that 
it enables learning a latent space of much higher dimensions.
Since the cosine similarity between two vectors lies in $[-1,1]$, 
assuming that $M_{ij} \in \left[ \alpha, \beta \right] $,
all entries in the original matrix $M$ 
are scaled during training according to
$M_{ij} = \dfrac{M_{ij} - \mu}{\mu - \alpha},  (i,j) \in \Omega,$
with $\mu = \left( \alpha + \beta \right) / 2$.

Employing \eqref{eq:reprLoss} to train the proposed model is equivalent to 
applying a \textit{manifold learning objective} on the intermediate outputs
of the row and column autoencoders.
The proposed model can then be thought of as performing two tasks simultaneously,
that is, \textit{learning to reconstruct} and \textit{manifold learning}.
The latter is the auxiliary task,
playing the role of improving the main reconstruction task.
\subsection{Objective Function}
While in single-task learning we optimize one loss function,
in multi-task learning, the objective function is a combination of loss terms.
In the proposed model, 
the loss associated to each of the known entries in $\Omega$ 
consists of three terms, namely, 
the reconstruction loss for the row autoencoder $\big(X_i(j) - \widehat{X}_i(j)\big)^2$, 
the reconstruction loss for the column autoencoder $\big(Y_j(i) - \widehat{Y}_j(i)\big)^2$,
and the representation loss defined in \eqref{eq:reprLoss},
where $X_i(j)$ denotes the $j$-th element of the $i$-th row,
and  $Y_j(i)$ the $i$-th element of the $j$-th column.
Averaging over all training samples in $\Omega$,
we obtain the objective function:
\begin{align}
\label{eq:proposedLoss}
        \mathcal{L} = \dfrac{1}{|\Omega|}  \sum_{ij \in \Omega} \Big[ \gamma_1 &\big(X_i(j) - \widehat{X}_i(j)\big)^2 + \gamma_2 \big(Y_j(i) - \widehat{Y}_j(i)\big)^2 \nonumber \\
        & + \gamma_3 \big( f(h_i, h_j) - M_{ij} \big)^2 \Big], 
\end{align}
with $\gamma_i > 0$, $i=\{1, 2, 3\}$,  appropriate weights.
\subsection{Comparison with Existing Work}
\label{sec:related work}
Multi-task learning has also been employed for MC in~\cite{dong17}. 
The model presented in~\cite{dong17} is based on 
a neural network component that incorporates side information,
coined additional stacked denoising autoencoder (aSDAE).
Specifically, aSDAE is a two-branch neural network that takes 
a pair of (corrupted) inputs, namely, 
a row (column) vector and the side information,
and produces a pair of reconstructed outputs.
Employing two branches that realize a row and a column aSDAE, 
the model performs matrix factorization as a main task,
using the latent representations provided by the row and column aSDAEs.
Concretely, the objective of this work concerns deep matrix factorization with side information.
Different from~\cite{dong17}, 
our work focuses on settings in which the available data is highly sparse
and side information is unavailable.
Our main objective is to compensate for the scarcity of data 
by providing an efficient regularization technique for autoencoder MC models
that can address overfitting.
%The proposed regularization is motivated by 
%and might apply to data that lie close to a manifold~\cite{Belkin06}.
In our setting, the row and column reconstruction is the main task
and the matrix factorization is an auxiliary task. 
Our approach is simpler with fewer branches compared to~\cite{dong17}  
and fewer hyperparameters to be tuned.

Other regularization methods aiming to improve the generalizability of AutoRec
have been reported in~\cite{strub16, strub16b, kuchaiev17}.
The model presented in~\cite{strub16, strub16b} 
uses a denoising autoencoder and employs side information to augment the training data 
and help the model learn better. 
The authors of~\cite{kuchaiev17} extend AutoRec with a very deep architecture 
and heavily regularize its parameters by employing dropout regularizer 
with very high dropping ratio,
while proposing a dense re-feeding technique to train the model. 
Nevertheless, the performance of these models is reduced 
in case of high scarcity in the training data or lack of side information. 
Similar to Dropout~\cite{Dropout}, 
denoising autoencoders average the realizations of a given ensemble
and try to make the model difficult to fit random error.
On the other hand, our model learns attributes that are present in the data
rather than preventing the model learning non-attributes.
It should be noted that existing techniques are complementary to the proposed approach
and their combination could lead to further performance improvement. 
%
%
%-----------------------------------------%
\section{Experimental Results}
\label{sec:experiment}
%-----------------------------------------%
We carry out experiments involving two benchmark datasets, 
namely, the MovieLens100K and the MovieLens1M~\cite{harper15}, 
containing approximately $100,000$ and one million observations, respectively. 
The datasets contain users' movie ratings summarized in two matrices 
where rows correspond to users and columns to movies.  
Only a small fraction of ratings is observed in each dataset. 
We randomly split these ratings into three sets; 
$75 \%$ are used for training, 
$5 \%$ for validation and $20 \%$ for evaluation. 
We evaluate the performance of the proposed model
in terms of regularization quality and reconstruction accuracy.
%and compare it against the AutoRec model proposed in~\cite{sedhain15}. 
On each dataset, we run the models on five different random splits of the data 
and report the average RMSE and MAE values,
calculated as follows: 
$\text{RMSE} = \sqrt{\sum_{ij \in \Omega_\text{eval}} (R_{ij} - M_{ij})^2 / \left| \Omega_\text{eval} \right|}$, and 
$\text{MAE} = \sum_{ij \in \Omega_\text{eval}} \left| R_{ij} - M_{ij} \right| / \left| \Omega_\text{eval} \right|$, 
%
%\begin{align*}
%    \text{RMSE} &= \sqrt{\sum_{ij \in \Omega_\text{eval}} (R_{ij} - M_{ij})^2 / \left| \Omega_\text{eval} \right|}, \\
%    \text{MAE}  &= \sum_{ij \in \Omega_\text{eval}} \left| R_{ij} - M_{ij} \right| / \left| \Omega_\text{eval} \right| ,
%    \label{eq:rmse_mae}
%\end{align*}
with $\Omega_\text{eval}$ the set of indices corresponding to entries available for evaluation 
and $\left| \Omega_\text{eval} \right|$ its cardinality.
\subsection{Hyperparameter Settings}
\label{sec:hyperparameters}
Following the original AutoRec model \cite{sedhain15}, 
we employ only one hidden layer with $500$ units for each branch 
of our model, while the numbers of units in the input and output layers are set 
according to the sizes of the matrices. 
We train our model using the Adam optimizer~\cite{adam}, 
with mini-batches of size $256$,
and initial learning rate equal to $0.01$ 
decaying by a factor $0.3$ every 30 epochs.
The number of training epochs is set to $200$. 
The $\ell_2$ regularization weight 
is set to $0.001$ for the MovieLens100K, 
and $0.00025$ for MovieLens1M dataset. 

We search for the best weights for the loss terms in \eqref{eq:proposedLoss}
using a separate random split of the two datasets. 
As the roles of row and column branches in our model are equivalent, 
we keep $\gamma_1$ and $\gamma_2$ fixed, equal to $1$, 
and only vary $\gamma_3$. 
The results for different values of $\gamma_3$ on the two datasets 
are shown in Table \ref{table:exp_gamma}.
For the MovieLens100K dataset, 
higher reconstruction accuracy is delivered with $\gamma_3=10.0$,
while for the MovieLens1M dataset, $\gamma_3=0.1$ is more effective.
The obtained values of $\gamma_3$ confirm our conjecture that 
the role of regularization is critical when
the number of available samples is small, 
that is, when we train our model on the MovieLens100K dataset.
\begin{table}[t]
\centering
\caption{Results (RMSE) of the proposed model on the MovieLens100K and MovieLens1M dataset when $\gamma_3$ varies.}
\label{table:exp_gamma}
\begin{tabular}{| c | c | c | c | c | c |}
\hline 
$\gamma_3$ & $0.1$ & $0.5$ & $1.0$ & $5.0$ & $10.0$  \\
\hhline {|=:=:=:=:=:=|}
& \multicolumn{5}{|c|}{MovieLens100K}  \\
\hhline {|=:=:=:=:=:=|}
Row-based & $0.963$ & $0.957$ & $0.953$ & $0.943$ & $\mathbf{0.944}$\\
%\hline 
%Row-based MAE & $0.763$ & $0.757$ & $0.754$ & $0.745$ & $0.746$\\
\hline
Column-based & $0.898$ & $0.898$ & $0.897$ & $0.894$ & $\mathbf{0.891}$\\
%\hline 
%Columnw-based MAE & $0.707$ & $0.707$ & $0.705$ & $0.704$ & $0.703$\\
\hhline {|=:=:=:=:=:=|}
& \multicolumn{5}{|c|}{MovieLens1M}  \\
\hhline {|=:=:=:=:=:=|}
Row-based & $\mathbf{0.876}$ & $0.877$ & $0.877$ & $0.876$ & $0.881$\\
%\hline 
%Row-based MAE & $0.763$ & $0.757$ & $0.754$ & $0.745$ & $0.746$\\
\hline
Column-based & $\mathbf{0.835}$ & $0.836$ & $0.836$ & $0.840$ & $0.847$\\
%\hline 
%Columnw-based MAE & $0.707$ & $0.707$ & $0.705$ & $0.704$ & $0.703$\\
\hline 
\end{tabular}
\end{table}
\subsection{Regularization Performance}
\begin{figure*}
    \centering
    \begin{subfigure}[t]{0.4\linewidth}
        \centering
        \includegraphics[width=0.8\linewidth]{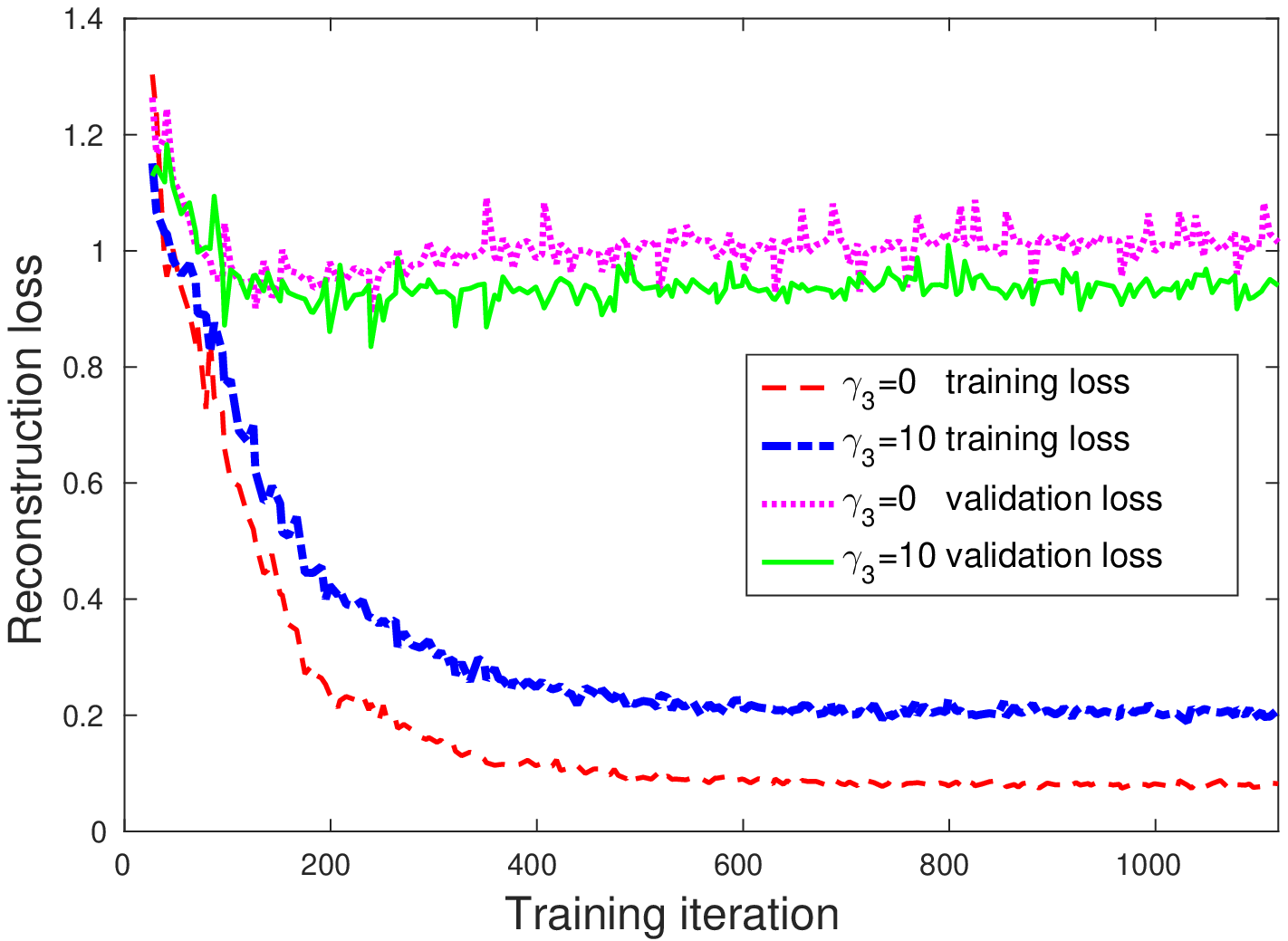}
        \caption{Without $\ell_2$-regularization}
        \label{fig:nol2}
    \end{subfigure}
    \begin{subfigure}[t]{0.4\linewidth}
        \centering
        \includegraphics[width=0.8\linewidth]{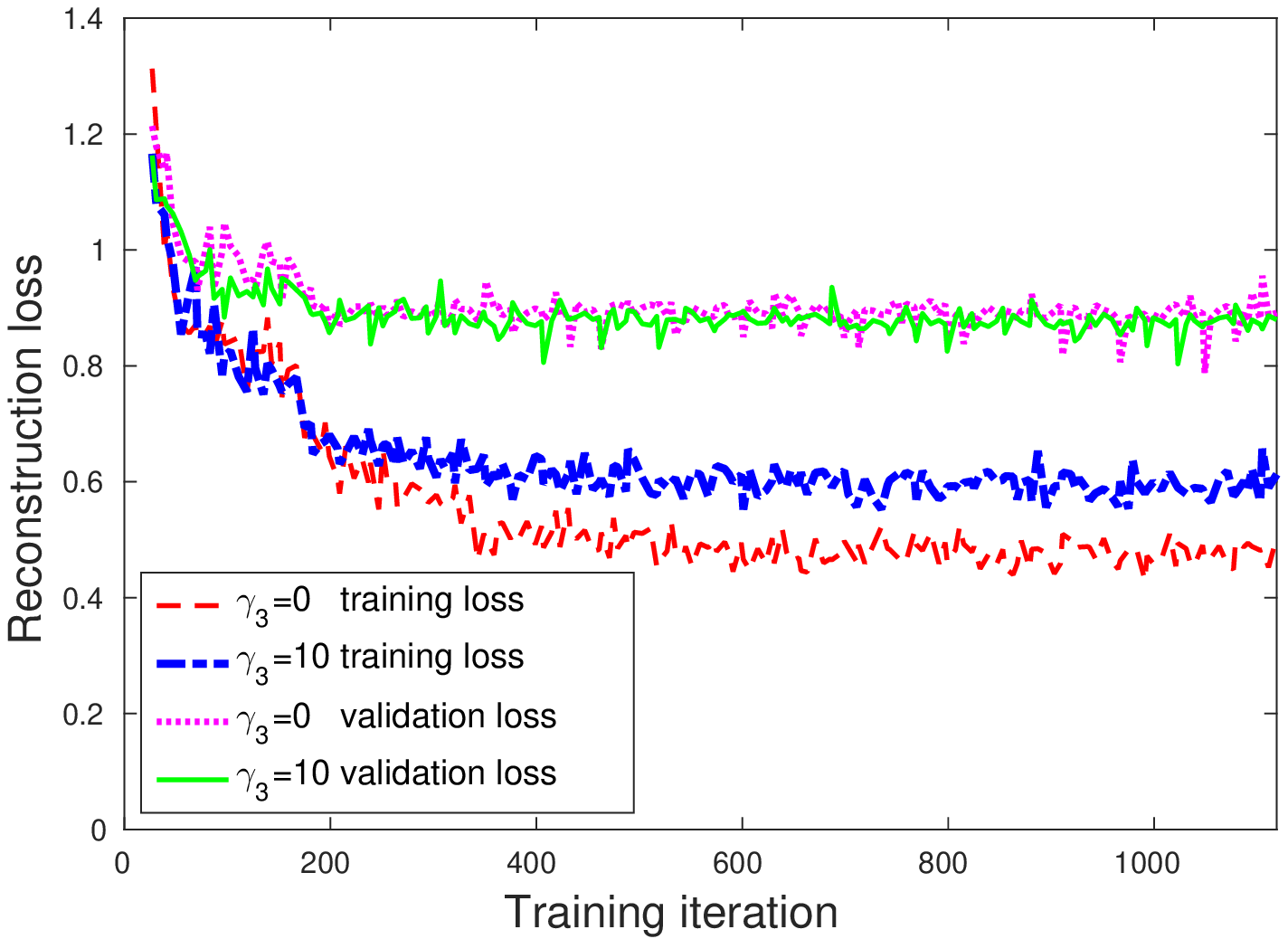}
        \caption{With $\ell_2$-regularization}
        \label{fig:l2}
    \end{subfigure}
    \caption{Evolution of training and validation reconstruction losses of the row-based Autorec model, on a split of the MovieLens100K dataset,
    under different regularization settings involving the proposed approach and the $\ell_2$ regularizer.}
    \label{fig:learning_curves}
\end{figure*}

As the proposed approach can be seen as 
a data-dependent regularizer, 
we carry out experiments on the MovieLens100K dataset 
to evaluate its regularization performance. 
Figure~\ref{fig:learning_curves} illustrates the training and validation reconstruction losses
of the proposed model while training proceeds. 
We train the model with and without $\ell_2$ regularization, 
for $\gamma_3=0$ and $\gamma_3=10$. 
When $\gamma_3=0$, the model reduces to
a row-based and a column-based Autorec working independently, 
and the proposed regularizer is not applied. 
As can be seen in Fig.~\ref{fig:nol2}, 
without any regularization, both models heavily overfit; 
however, when the proposed approach is applied ($\gamma_3 = 10$),
the generalization error (i.e., the gap between the training and the validation loss) becomes lower. 
The generalization errors decrease further when the $\ell_2$ regularization is employed (Fig.~\ref{fig:l2}).
Combining both the $\ell_2$ regularization and the proposed approach,
we obtain the best results in terms of generalization error.
This confirms that different regularization techniques can be complementary to each other. 

We note that for a highly tuned problem like matrix completion, 
a small improvement in the validation loss
that improves the reconstruction accuracy 
is significant (see also Sec.~\ref{sec:recon_accuracy}). 
\subsection{Reconstruction accuracy}
\label{sec:recon_accuracy}
We carry out experiments to compare the proposed model 
to other neural-network-based MC models in terms of reconstruction quality. 
In this experiment, we use the tuned values of $\gamma_3$ (see Sec.~\ref{sec:hyperparameters})
while all other hyperparameter settings are the same.  
Table \ref{table:final_results} presents a comparison between the proposed model 
against the AutoRec model~\cite{sedhain15} and the matrix factorization model proposed in~\cite{nguyen18}. 
We do not include a comparison with the more recent autoencoder-based models~\cite{strub16,strub16b}, 
as they do not improve over AutoRec when the training data is small, which is the focus of this work.
\begin{table}
\centering
\caption{Comparison against neural-network-based models on the MovieLens100K and MovieLens1M datasets.}
\label{table:final_results}
\begin{tabular}{| c | c | c | c | c |}
\hline
\multirow{2}{*}{}&\multicolumn{2}{|c|}{MovieLens100K} & \multicolumn{2}{|c|}{MovieLens1M} \\
\hhline {|~|=:=:=:=|}
& RMSE & MAE & RMSE & MAE  \\
\hhline {|=:=:=:=:=|}
Nguyen et al. \cite{nguyen18} & $0.895$ & $0.703$ & $0.849$ & $0.673$ \\
\hline
AutoRec \cite{sedhain15} (row-based) & $0.962$ & $0.764$ & $0.878$ & $0.699$  \\
\hline 
Proposed (row-based) & $0.937$ & $0.779$ & $0.877$ & $0.698$  \\
\hline
AutoRec \cite{sedhain15} (column-based) & $0.898$ & $0.705$ & $0.836$ & $0.657$ \\
\hline 
Proposed (column-based) & $\mathbf{0.890}$ & $\mathbf{0.700}$ & $\mathbf{0.835}$ & $\mathbf{0.656}$ \\
\hline
\end{tabular}
\end{table}
As can be seen, the proposed model consistently improves over 
the corresponding Autorec models (row-based and column-based). 
It should be noted on MovieLens datasets, column-based Autorec usually performs better 
than row-based Autorec \cite{sedhain15} as the number of ratings per item is usually higher than per user. 
The improvements are more significant on the MovieLens100K dataset, 
which has far fewer training data than the MovieLens1M dataset. 
On the MovieLens100K dataset, 
the column branch of the proposed model outperforms \cite{nguyen18} 
which generalizes much better than the AutoRec model. 
On the MovieLens1M dataset, 
the column branch of our model delivers the best results, 
followed by the column-based AutoRec and \cite{nguyen18}. 
\section{Conclusion}
\label{sec:conclusion}
We propose a data-dependent regularizer for autoencoder-based matrix completion models. 
Our approach relies on the multi-task learning principle, 
performing prediction as the main task and manifold learning as an auxiliary task.   
The latter acts as a regularizer, improving the generalizability on the main task. 
Experimental results on two real-world datasets show that 
the proposed approach effectively reduces overfitting 
for both row and column autoencoder-based models, 
especially, when the number of training data is small; 
and consistently outperforms state-of-the-art models in terms of reconstruction accuracy.
\bibliographystyle{IEEEtran}
\bibliography{autorecRefs}

% Generated by IEEEtran.bst, version: 1.14 (2015/08/26)
\begin{thebibliography}{10}
\providecommand{\url}[1]{#1}
\csname url@samestyle\endcsname
\providecommand{\newblock}{\relax}
\providecommand{\bibinfo}[2]{#2}
\providecommand{\BIBentrySTDinterwordspacing}{\spaceskip=0pt\relax}
\providecommand{\BIBentryALTinterwordstretchfactor}{4}
\providecommand{\BIBentryALTinterwordspacing}{\spaceskip=\fontdimen2\font plus
\BIBentryALTinterwordstretchfactor\fontdimen3\font minus
  \fontdimen4\font\relax}
\providecommand{\BIBforeignlanguage}[2]{{%
\expandafter\ifx\csname l@#1\endcsname\relax
\typeout{** WARNING: IEEEtran.bst: No hyphenation pattern has been}%
\typeout{** loaded for the language `#1'. Using the pattern for}%
\typeout{** the default language instead.}%
\else
\language=\csname l@#1\endcsname
\fi
#2}}
\providecommand{\BIBdecl}{\relax}
\BIBdecl

\bibitem{Cao2014}
F.~Cao, M.~Cai, and Y.~Tan, ``Image interpolation via low-rank matrix
  completion and recovery,'' \emph{IEEE Trans. Circuits Syst. Video Technol.},
  vol.~25, no.~8, pp. 1261 -- 1270, 2015.

\bibitem{Ji2010}
H.~Ji, C.~Liu, Z.~Shen, and Y.~Xu, ``Robust video denoising using low rank
  matrix completion,'' \emph{IEEE Conf.~Computer Vision and Pattern Recognition
  (CVPR)}, pp. 1791--1798, 2010.

\bibitem{zheng16}
Y.~Zheng, B.~Tang, W.~Ding, and H.~Zhou, ``A neural autoregressive approach to
  collaborative filtering,'' in \emph{Int.~Conf.~Machine Learning (ICML)},
  2016, pp. 764--773.

\bibitem{monti17}
F.~Monti, M.~M. Bronstein, and X.~Bresson, ``Geometric matrix completion with
  recurrent multi-graph neural networks,'' \emph{arXiv:1704.06803}, 2017.

\bibitem{candes09}
E.~J. Cand{\`e}s and B.~Recht, ``Exact matrix completion via convex
  optimization,'' \emph{Found.~Comput.~Math.}, vol.~9, no.~6, p. 717, 2009.

\bibitem{Recht10}
B.~Recht, M.~Fazel, and P.~A. Parrilo, ``Guaranteed minimum-rank solutions of
  linear matrix equations via nuclear norm minimization,'' \emph{{SIAM}
  Review}, vol.~52, no.~3, pp. 471--501, 2010.

\bibitem{salakhutdinov07}
R.~Salakhutdinov and A.~Mnih, ``Probabilistic matrix factorization,'' in
  \emph{Adv.~Neural Inf.~Process.~Syst.~(NIPS)}, 2007, pp. 1257--1264.

\bibitem{Koren2009}
Y.~Koren, R.~Bell, and C.~Volinsky, ``{M}atrix {F}actorization {T}echniques for
  {R}ecommender {S}ystems,'' \emph{IEEE Computer}, vol.~42, no.~8, pp. 30--37,
  2009.

\bibitem{volkovs17}
M.~Volkovs, G.~Yu, and T.~Poutanen, ``Dropoutnet: Addressing cold start in
  recommender systems,'' in \emph{Adv.~Neural Inf.~Process.~Syst.~(NIPS)},
  2017, pp. 4957--4966.

\bibitem{nguyen18}
D.~M. Nguyen, E.~Tsiligianni, and N.~Deligiannis, ``Extendable neural matrix
  completion,'' in \emph{IEEE Int.~Conf.~Acoust.~Speech Signal
  Process.~(ICASSP) [Available: arXiv:1805.04912]}, 2018.

\bibitem{nguyen18_spl}
------, ``Learning discrete matrix factorization models,'' \emph{IEEE Signal
  Process. Lett.}, vol.~25, no.~5, pp. 720--724, 2018.

\bibitem{mohamed12}
A.~r.~Mohamed, G.~E. Dahl, and G.~Hinton, ``Acoustic modeling using deep belief
  networks,'' \emph{IEEE Trans. Audio, Speech, Language Process.}, vol.~20,
  no.~1, pp. 14--22, 2012.

\bibitem{nguyen17}
D.~M. Nguyen, E.~Tsiligianni, and N.~Deligiannis, ``Deep learning sparse
  ternary projections for compressed sensing of images,'' in \emph{IEEE Global
  Conference on Signal and Information Processing (GlobalSIP)}, 2017, pp.
  1125--1129.

\bibitem{he16}
K.~He, X.~Zhang, S.~Ren, and J.~Sun, ``Deep residual learning for image
  recognition,'' in \emph{IEEE Conf.~Computer Vision and Pattern Recognition
  (CVPR)}, 2016, pp. 770--778.

\bibitem{do18}
T.~H. Do, D.~M. Nguyen, E.~Tsiligianni, B.~Cornelis, and N.~Deligiannis,
  ``Twitter user geolocation using deep multiview learning,'' in \emph{IEEE
  Int.~Conf.~Acoust.~Speech Signal Process.~(ICASSP) [Available:
  arXiv:1805.04612]}, 2018.

\bibitem{sedhain15}
S.~Sedhain, A.~K. Menon, S.~Sanner, and L.~Xie, ``{AutoRec: Autoencoders Meet
  Collaborative Filtering},'' in \emph{Int.~Conf.~World Wide Web (WWW)}, 2015,
  pp. 111--112.

\bibitem{strub16}
F.~Strub, R.~Gaudel, and J.~Mary, ``Hybrid recommender system based on
  autoencoders,'' in \emph{The 1st Workshop on Deep Learning for Recommender
  Systems (DLRS)}, 2016, pp. 11--16.

\bibitem{strub16b}
F.~Strub, J.~Mary, and R.~Gaudel, ``Hybrid collaborative filtering with neural
  networks,'' \emph{arXiv:1603.00806}, 2016.

\bibitem{dong17}
X.~Dong, L.~Yu, Z.~Wu, Y.~Sun, L.~Yuan, and F.~Zhang, ``A hybrid collaborative
  filtering model with deep structure for recommender systems,'' in \emph{AAAI
  Conf.~Artificial Intelligence}, 2017, pp. 1309--1315.

\bibitem{kuchaiev17}
O.~Kuchaiev and B.~Ginsburg, ``Training deep autoencoders for collaborative
  filtering,'' \emph{arXiv:1708.01715}, 2017.

\bibitem{Dropout}
N.~Srivastava, G.~Hinton, A.~Krizhevsky, I.~Sutskever, and R.~Salakhutdinov,
  ``Dropout: A simple way to prevent neural networks from overfitting,''
  \emph{J.~Mach.~Learn.~Res.}, vol.~15, pp. 1929--1958, 2014.

\bibitem{Belkin06}
M.~Belkin, P.~Niyogi, and V.~Sindhwani, ``Manifold regularization: {A}
  geometric framework for learning from labeled and unlabeled examples,''
  \emph{J.~Mach.~Learn.~Res.}, vol.~7, pp. 2399--2434, 2006.

\bibitem{GraphConnect}
J.~Huang, Q.~Qiu, R.~Calderbank, and G.~Sapiro, ``{GraphConnect}: {A}
  regularization framework for neural networks,'' \emph{arXiv:1512.06757},
  2015.

\bibitem{Caruana1997}
R.~Caruana, ``Multitask learning,'' \emph{Machine Learning}, vol.~28, no.~1,
  pp. 41--75, Jul. 1997.

\bibitem{Evgeniou05}
T.~Evgeniou, C.~A. Micchelli, and M.~Pontil, ``Learning multiple tasks with
  kernel methods,'' \emph{J.~Mach.~Learn.~Res.}, vol.~6, pp. 615--637, 2005.

\bibitem{ruder17}
S.~Ruder, ``An overview of multi-task learning in deep neural networks,''
  \emph{arXiv:1706.05098}, 2017.

\bibitem{Caruana93}
R.~Caruana, ``{Multitask Learning: A Knowledge-Based Source of Inductive
  Bias},'' in \emph{Int.~Conf.~Machine Learning}.\hskip 1em plus 0.5em minus
  0.4em\relax Morgan Kaufmann, 1993, pp. 41--48.

\bibitem{Baxter2000}
J.~Baxter, ``A model of inductive bias learning,'' \emph{Journal of Artificial
  Intelligence Research}, vol.~12, no.~1, pp. 149--198, Mar. 2000.

\bibitem{BenDavid2003}
S.~Ben-David and R.~Schuller, ``Exploiting task relatedness for multiple task
  learning,'' in \emph{Learning Theory and Kernel Machines}, 2003, pp.
  567--580.

\bibitem{harper15}
F.~M. Harper and J.~A. Konstan, ``{The MovieLens Datasets: History and
  Context},'' \emph{ACM Trans. Interact. Intell. Syst.}, vol.~5, no.~4, pp.
  19:1--19:19, 2015.

\bibitem{adam}
D.~P. Kingma and J.~Ba, ``Adam: {A} method for stochastic optimization,''
  \emph{arXiv:1412.6980}, 2014.

\end{thebibliography}
\end{document}